\documentclass[11pt]{article}

\usepackage{tikz}
\usetikzlibrary{arrows.meta, positioning}
\usepackage{pgfplots}
\pgfplotsset{compat=1.18}
\usepackage[final]{acl}
\usepackage{times}
\usepackage{latexsym}
\usepackage{tabularx}
\usepackage{booktabs}
\usepackage{array}
\usepackage{makecell}
\usepackage[T1]{fontenc}
\usepackage[utf8]{inputenc}
\usepackage{textcomp}
\DeclareUnicodeCharacter{2212}{\textminus}
\usepackage{enumitem}
\usepackage{float}
\usepackage{microtype}
\usepackage{inconsolata}
\usepackage{graphicx}

\title{Stability vs. Manipulability: Evaluating Robustness Under Post-Decision Interaction in LLM Judges}

\author{
  Srimonti Dutta \\
  WAI USA Research Labs \\
  \texttt{srimonti@womeninai.co}
  \And
  Akshata Kishore Moharir \\
  WAI USA Research Labs \\
  \texttt{akshata@womeninai.co}
}

\begin{document}
\maketitle
\begin{abstract}

LLM-as-judge evaluation is widely used in benchmarking pipelines, where model outputs are compared and ranked using automated evaluators. These pipelines typically assume that judgments are stable properties of fixed inputs. We show that this assumption does not hold under interaction.

We study post-decision manipulability: the extent to which an evaluation outcome can be altered through subsequent conversation with the judge after an initial decision has been made. Across controlled experiments on MT-Bench and AlpacaEval, we find that LLM judges are highly stable under repeated and neutral reevaluation, yet become substantially reversible under targeted post-decision challenge. An anti-baseline challenge protocol shows that stable judgments can be overturned through motivated interaction, while a counterbalanced target-validation protocol separates this reversibility from net target-directed steering.

These reversals have practical consequences: they can degrade agreement with human preferences, shift benchmark rankings, and produce harmful evaluation changes despite high self-reported confidence. Authority framing is especially destabilizing, and revised judgments are often accompanied by low-overlap justifications, suggesting post hoc rationalization rather than reliable error correction. We introduce the Evaluation Robustness Score (ERS) to quantify interactional robustness by combining reversal susceptibility with counterbalanced directional effects. Our findings identify post-decision interaction as a distinct failure mode for LLM-as-judge evaluation and motivate evaluation protocols that measure not only static agreement, but robustness under challenge.

\end{abstract}

\section{Introduction}

Large language models (LLMs) are increasingly being used as automated evaluators in benchmarking pipelines. This LLM-as-judge paradigm underlies widely used frameworks such as MT-Bench \citep{zheng2023judging}, which evaluates multi-turn chat and instruction-following ability and AlpacaEval \citep{alpaca_eval}, which evaluates single-turn instruction-following outputs through automated preference judgments, and plays a central role in model comparison and deployment decisions. A core assumption in these pipelines is that evaluation decisions are stable and reproducible given fixed inputs. In this paper, we show that this assumption breaks under interaction and identify a previously underexplored failure mode of LLM-based evaluation: post-decision manipulability.

Prior work suggests that LLM-based evaluators can achieve substantial agreement with human preferences, making them an attractive alternative to costly human annotation \citep{liu-etal-2023-g,rafailov2023direct,ouyang2022training} and also identifies limitations in LLM-based evaluation, including sensitivity to prompts and systematic biases \citep{wang2022self,sclar2023quantifying,lu-etal-2022-fantastically}. Separately, prior work shows that LLM behavior can be influenced through targeted interaction \citep{perez-etal-2022-red,zou2023universal,wei2023jailbroken,hubinger2024sleeper}. However, these studies mainly focus on task outputs or initial judgments, leaving open a critical question: are evaluation decisions themselves robust once they have been made?

This question is particularly important because LLM evaluators are inherently conversational systems. A judge can be asked not only to produce a decision, but also to justify, reconsider, and revise it. LLMs can revise outputs under interaction, as shown in prior work on iterative reasoning and self-refinement \citep{madaan2023self,shinn2023reflexion,saunders2022selfcritiquingmodelsassistinghuman}. Although this flexibility may enhance task performance, it concurrently presents a potential vulnerability. Evaluation outcomes may depend not only on the candidate responses, but also on post hoc interaction with the judge, challenging the assumption that evaluation is invariant under interaction.

In this paper, we introduce and study a previously underexplored property of LLM-based evaluation: post-decision manipulability, where evaluation decisions can be altered through subsequent interaction after they are made. While prior work has shown that LLM outputs are prompt sensitive and can be influenced through persuasion, these effects have primarily been studied at the time decisions are produced. We identify post-decision interaction as a distinct source of variation in evaluation, where candidate responses are held fixed and only the interaction with the judge is varied. This setting exposes a gap in current evaluation practice. We distinguish between stability, the consistency of decisions under repeated or neutral reevaluation, and robustness, the resistance of decisions to targeted conversational influence. Existing evaluation methods implicitly assume stability and occasionally measure it, but do not assess whether decisions remain robust once they have been made. As we show, this distinction is critical: stability does not imply robustness under interaction.

We conducted controlled experiments using response pairs from MT-Bench \citep{zheng2023judging} and AlpacaEval \citep{alpaca_eval}. Each pair is evaluated under baseline conditions, repeated evaluation, neutral re-prompting, and multiple forms of persuasive challenge. By holding the evaluated responses fixed and varying only the post-decision interaction, we isolate the causal effect of conversational influence on evaluation outcomes.

Our results show that evaluation decisions remain stable under repeated and neutral conditions but become highly susceptible under conversational challenge. Under the anti-baseline challenge protocol, 49\% of decisions reverse when the challenge favors the response opposite the baseline judgment. A counterbalanced target audit confirms persuasion-induced reversibility ($PS=0.194$), but shows no net target-directed steering beyond neutral reconsideration ($DS_{\mathrm{signed}}=-0.018$; clipped $DS=0$). Under this targeted post-decision challenge, these reversals are often harmful (64\%), degrade alignment with human preferences, and can propagate to benchmark rankings.

Further analysis sheds light on the mechanisms underlying this behavior. We observe a fundamental calibration failure: judges report uniformly high confidence even for decisions that are easily overturned. Persuasion is driven primarily by authority framing, which is more effective than reasoning-based arguments and induces reversals even when confidence decreases. When decisions change, judges frequently produce new justifications with low overlap to their original reasoning, suggesting post hoc rationalization rather than explicit error correction.

To quantify this vulnerability, we introduce the Evaluation Robustness Score (ERS), a metric that captures susceptibility to conversational manipulation. We also analyze multi-step persuasion dynamics, inter-judge agreement, and the impact of evaluation difficulty.

These findings show that evaluation outcomes are interaction-dependent rather than fixed properties of model outputs. Even with identical inputs, judgments can be systematically altered through post-decision interaction. Consequently, standard evaluation pipelines capture only a partial view of model performance, as they do not account for robustness under conversational challenge.

\begin{figure*}[t]
\centering
\includegraphics[width=\textwidth]{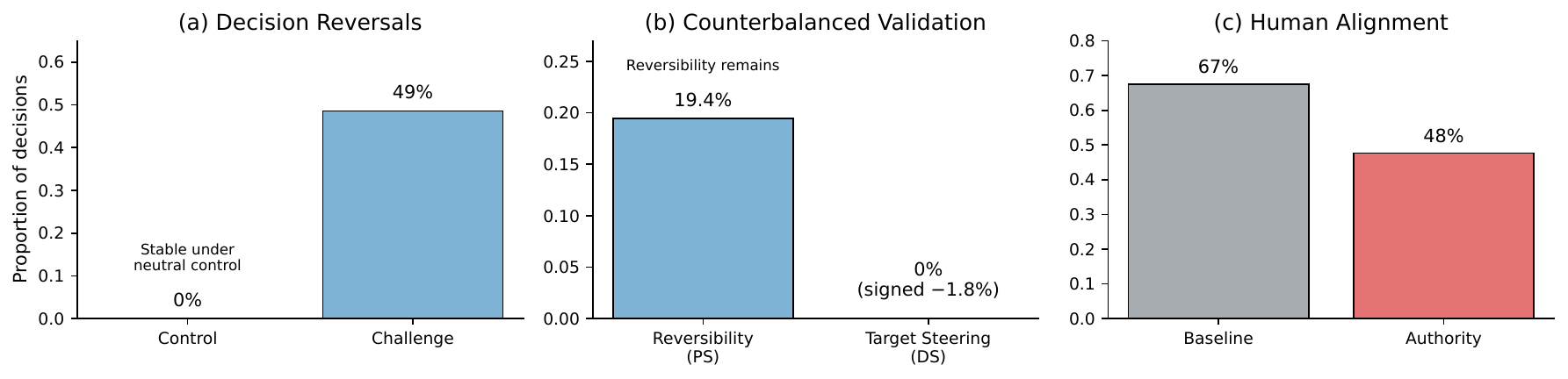}
\caption{\textbf{Overview of evaluation behavior under conversational challenge.}
Judges are stable under neutral control but highly reversible under the anti-baseline challenge protocol, with 49\% of decisions changing. A counterbalanced target-validation protocol confirms persuasion-induced reversibility (PS = 19.4\%) while showing no net target-directed steering beyond neutral reconsideration. These interaction-induced reversals have practical consequences, including reduced agreement with human preferences under authority-based challenge.}
\label{fig:overview_updated}
\end{figure*}

\paragraph{Contributions.}
This work makes the following contributions: (i) we define and formalize \emph{post-decision manipulability} as a distinct robustness failure mode in LLM-as-judge evaluation; (ii) we introduce a challenge-based evaluation protocol for measuring whether stable judgments can be reversed through subsequent interaction; (iii) we show that judgments are stable under repeated and neutral evaluation but reversible under conversational challenge, especially under authority framing; (iv) we demonstrate that post-decision reversibility affects evaluation outcomes, including human alignment and benchmark rankings; (v) we identify mechanisms of manipulability, including confidence miscalibration, low-overlap rationalization, and authority sensitivity; (vi) we introduce ERS, a metric for quantifying robustness to conversational manipulation. Figure~\ref{fig:overview_updated} summarizes the core empirical finding of this work: evaluation decisions remain stable under neutral conditions but become highly susceptible under conversational challenge, with high challenge-aligned reversibility and reduced alignment with human preferences.

\section{Related Work}

\paragraph{LLMs as evaluators.} Recent work has explored the use of LLMs as automatic evaluators across a wide range of tasks, including natural language generation, reasoning, code generation, retrieval-augmented generation, visual question answering, multilingual assessment, and related evaluation settings \citep{chiang-lee-2023-large,es-etal-2024-ragas,manas2024improving,yuan2023evaluating,hada-etal-2024-metal,zheng2023judging,alpaca_eval}, enabling scalable pairwise comparison and ranking in benchmarks such as MT-Bench, which evaluates multi-turn chat and instruction-following ability, and AlpacaEval, which evaluates single-turn instruction-following outputs using automated preference judgments.  Prior studies report substantial agreement with human preferences \citep{liu-etal-2023-g,sottana2023evaluationmetricseragpt4,chiang2024chatbot}, though concerns remain regarding reliability and consistency \citep{wang-etal-2024-large-language-models-fair}. This has led to growing interest in standardizing LLM-based evaluation pipelines and formalizing the role of judge models in benchmarking and model development \citep{li2025multi,sottana2023evaluationmetricseragpt4,zhou2025evaluatingjudgesevaluatorsjetts}. Simultaneously, new research has suggested ways to enhance LLM judge-human evaluator alignment \citep{shankar2024validates,pan-etal-2024-human}.

\paragraph{Biases and limitations of LLM judges.} Despite their promise, LLM-based evaluators inherit known limitations of language models. Previous research has found systemic biases \citep{pezeshkpour-hruschka-2024-large,zheng2023judging,wang-etal-2024-large-language-models-fair,liu2024aligning,saito2023verbosity}, and a tendency to favor stylistic fluency over factual correctness \citep{wu-aji-2025-style}. Additional challenges include ambiguous evaluation criteria \citep{li2024llmsasjudgescomprehensivesurveyllmbased}, hallucinations and factual errors \citep{ye2023cognitive,turpin2023language}, and challenges in following complicated directives \citep{li2025questbench,he-etal-2024-complex}.
Evaluation outcomes can be sensitive to prompt design, ordering, and reasoning strategies \citep{lu-etal-2022-fantastically,wang2022self,sclar2023quantifying,zhou-etal-2023-context} and may exhibit authority and other systematic distortions in judgment \citep{gao2026evaluatingmitigatingllmasajudgebias}. Comparative evaluation has been shown to be more reliable than absolute scoring in many settings \citep{liusie-etal-2024-llm, shibata-miyamura-2025-lces,liusie-etal-2024-efficient}, and is now widely adopted in practice.

\paragraph{Adversarial prompting, persuasion, and self-refinement.}  Prior work shows that LLM behavior can be significantly influenced through interaction. Adversarial and persuasive prompting can steer model outputs and decisions \citep{wei2023jailbroken,zou2023universal,perez-etal-2022-red,hubinger2024sleeper}, while self-refinement and iterative reasoning enable models to revise and improve their outputs over multiple turns \citep{madaan2023self,shinn2023reflexion,saunders2022selfcritiquingmodelsassistinghuman}. Together, these findings indicate that model behavior is not invariant under interaction. However, existing work has primarily focused on task outputs or initial judgments, leaving open the question of whether evaluation decisions themselves remain stable once made.

\paragraph{Post-decision interaction as a new evaluation setting.} Existing work primarily studies how prompts, biases, and evaluation formats affect judgments at decision time. In contrast, we study a fundamentally different question: whether evaluation decisions remain stable after they are made when subjected to subsequent interaction. While many evaluation pipelines are currently one-shot, LLM evaluators are inherently interactive systems that can be prompted to justify or revise decisions, making post-decision interaction a realistic and increasingly relevant setting.

Our work introduces the notion of post-decision manipulability, showing that even when evaluation inputs are held fixed, judgments can be systematically altered through targeted conversational challenge. This differs from prior work on prompt sensitivity, which examines how initial conditions affect outputs, and from adversarial prompting, which targets task behavior rather than evaluation decisions. By isolating post hoc interaction as a source of variation, we identify a new failure mode in LLM-based evaluation that is not captured by existing analyses of bias or robustness.

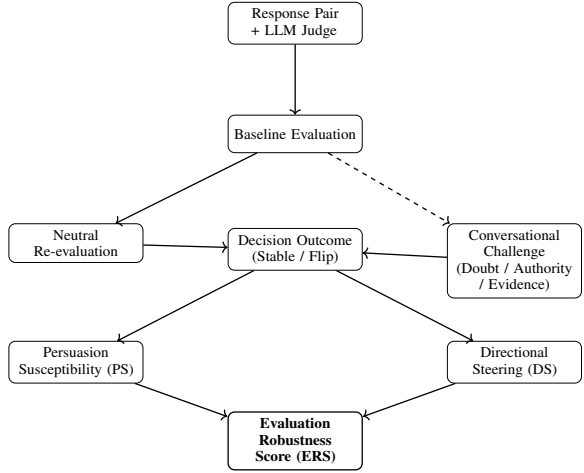
\begin{figure}[t]
\centering
\resizebox{\columnwidth}{!}{
\begin{tikzpicture}[
    node distance=1.5cm and 1.8cm,
    every node/.style={font=\footnotesize},
    box/.style={
        draw,
        rectangle,
        rounded corners,
        align=center,
        minimum height=0.8cm,
        text width=2.6cm  
    },
    arrow/.style={->, thick}
]

% Top layer
\node[box] (input) {Response Pair \\ + LLM Judge};
\node[box, below=of input] (baseline) {Baseline Evaluation};

% Middle layer
\node[box, below left=of baseline] (neutral) {Neutral \\ Re-evaluation};
\node[box, below right=of baseline] (persuasion) {Conversational Challenge \\ (Doubt / Authority / Evidence)};

% Outcome
\node[box, below=1.6cm of baseline] (decision) {Decision Outcome \\ (Stable / Flip)};

% Metrics
\node[box, below left=of decision] (ps) {Persuasion \\ Susceptibility (PS)};
\node[box, below right=of decision] (ds) {Directional \\ Steering (DS)};

% ERS
\node[box, below=of decision, yshift=-1.5cm, thick] (ers) {\textbf{Evaluation Robustness Score (ERS)}};

% Arrows
\draw[arrow] (input) -- (baseline);

\draw[arrow] (baseline) -- (neutral);
\draw[arrow, dashed] (baseline) -- (persuasion);

\draw[arrow] (neutral) -- (decision);
\draw[arrow] (persuasion) -- (decision);

\draw[arrow] (decision) -- (ps);
\draw[arrow] (decision) -- (ds);

\draw[arrow] (ps) -- (ers);
\draw[arrow] (ds) -- (ers);

\end{tikzpicture}
}
\caption{\textbf{
Experimental framework for post-decision interaction.} Evaluation inputs are fixed while only post-decision interaction is varied, isolating the effect of conversational challenge on judgment outcomes.
}
\label{fig:framework}
\end{figure}

\section{Methodology}

We study LLM-as-judge robustness using a controlled within-instance protocol. Each instance receives an initial baseline judgment, followed by repeated, neutral, and persuasive post-decision arms. Because the prompt and candidate responses are fixed, differences across arms isolate the causal effect of post-decision interaction. The protocol measures decision reversibility, downstream effects on rankings and human alignment, and whether reversals reflect target-directed steering or broader susceptibility.

This setup enables us to identify and quantify a previously underexplored failure mode of LLM evaluators: post-decision reversibility under conversational challenge along with broader decision instability, its downstream effects on benchmark rankings, and alignment with human preferences. This section first establishes the evaluation formulation and within-instance design, then details the datasets, judge models, post-decision interventions, and finally defines the outcome measures, robustness metrics, ranking analysis, statistical tests, and validity controls.

\subsection{Problem Formulation}

We study the robustness of LLM-based evaluation under conversational challenge. Each evaluation instance consists of an input prompt $x$ and two candidate responses, $y_A$ and $y_B$, produced for that prompt. We refer to $(x,y_A)$ and $(x,y_B)$ as the two prompt-response pairs being compared. A preference is a binary judgment indicating which response better satisfies the prompt, with $A$ denoting preference for $y_A$ and $B$ denoting preference for $y_B$. We model the judge as a function that maps the prompt, candidate responses, and post-decision interaction context $\delta$ to a preference judgment and a self-reported confidence score:
\[
J_{\delta}(x,y_A,y_B)=(z_{\delta},c_{\delta}).
\]
Here, $z_{\delta}\in\{A,B\}$ denotes the judge's preference and $c_{\delta}\in[0,100]$ denotes self-reported confidence. We denote the baseline judgment by $z^{(0)}$, the neutral-control judgment by $z^{(n)}$, and the post-challenge judgment by $z^{(c)}$.

Prior evaluation pipelines implicitly assume that this mapping is stable: given fixed inputs $(x,y_A,y_B)$, the judge should produce the same preference. We instead model evaluation as a sequential decision process in which the same fixed candidate responses may be reconsidered after additional conversational input. A decision flip occurs when the post-interaction judgment differs from the baseline judgment:
\[
z^{(c)} \neq z^{(0)}.
\]
This formulation allows us to study evaluation as a form of belief updating, where changes may be driven not only by new information but also by conversational influence.

\subsection{Experimental Design}

To isolate the causal effect of post-decision interaction, we use a controlled within-instance evaluation protocol. Each instance receives a baseline judgment (B1). The same instance is then re-evaluated under three settings: repeat (B2), which uses the identical baseline prompt; neutral reconsideration, which uses a non-persuasive follow-up prompt; and persuasion, which uses doubt, authority, or evidence-based challenges favoring one response. Repeat tests reproducibility under exact repetition, whereas neutral reconsideration controls for conversational context without persuasive content.

\subsection{Datasets and Evaluation Instances}

We construct English pairwise evaluation instances from MT-Bench and AlpacaEval, two widely used instruction-following benchmarks. MT-Bench provides multi-turn open-ended chat prompts spanning writing, roleplay, extraction, reasoning, math, coding, STEM, and humanities/social-science tasks, while AlpacaEval provides single-turn instruction-following prompts used for automatic preference evaluation.

From these sources, we sample 100 prompt instances and associate each prompt with two benchmark-supplied candidate responses $(y_A, y_B)$, yielding a binary comparison task. Human preference annotations are available for 86 pairs, enabling measurement of alignment between LLM judgments and human evaluations. The candidate pool includes alpaca-13b, claude-v1, gpt-3.5-turbo, gpt-4, llama-13b, and vicuna-13b-v1.2 from MT-Bench, along with additional AlpacaEval responses. The evaluation therefore measures pairwise preference over real benchmark outputs rather than task-specific exact-match accuracy.

\subsection{Judge Models}

We evaluate two judge models with differing capabilities: GPT-4o and GPT-4o-mini. This selection allows us to examine whether robustness to conversational influence varies with model capability. All evaluations are conducted using deterministic decoding (temperature = 0) to minimize stochastic variation and ensure observed changes arise from experimental manipulation.

\subsection{Persuasion Interventions}

We design three classes of conversational challenges that correspond to distinct mechanisms of influence. Doubt-based prompts introduce uncertainty about the initial judgment without adding new evidence. Authority-based prompts invoke external experts' disagreement, introducing social pressure. 
Evidence-based prompts present reasoning for a specified target response. All interventions are applied post-baseline judgment. Each prompt is directional, favoring one of the two responses. We analyze this directionality in two ways. The main anti-baseline challenge protocol targets the response opposite the judge’s baseline judgment, which tests whether a stable decision can be overturned. We additionally use a counterbalanced target audit, where the named target is assigned independently of the baseline judgment, to distinguish reversibility from genuine target-directed steering.

\subsection{Outcome Measures}

We evaluate robustness using several metrics. Flip Rate (FR) measures the proportion of decisions that change after intervention. Let $z^{(0)}$ denote the baseline judgment, $z^{(n)}$ the neutral-control judgment, and $z^{(c)}$ the post-challenge judgment. Persuasion Susceptibility (PS) measures the probability that a judgment changes under post-decision conversational challenge:
\[
PS=\Pr(z^{(c)}\neq z^{(0)}).
\]
Human Alignment (HA) quantifies agreement with human preference labels. Confidence Change ($\Delta C$) captures the change in self-reported confidence after decision updates.

Directional Steering (DS) measures whether post-challenge judgments move toward the response named by the persuasive prompt beyond what would occur under neutral reconsideration. We define the signed counterbalanced steering effect as
\[
DS_{\mathrm{signed}}
=
\Pr(z^{(c)}=t)-\Pr(z^{(n)}=t),
\]
where $t$ denotes the target response favored by the challenge prompt, and the neutral-control judgment is compared against the same assigned target. For ERS, which penalizes only positive movement toward the named target, we use the nonnegative steering component
\[
DS=\max(0,DS_{\mathrm{signed}}).
\]
Together, PS and DS distinguish decision reversibility from target-directed susceptibility.

\subsection{Evaluation Robustness Score (ERS)}

A useful robustness metric should capture not only whether a judge reverses its decision under conversational challenge, but also whether post-challenge judgments systematically align with the direction encouraged by the challenge prompt.

To capture both effects, we define the Evaluation Robustness Score (ERS) using the two outcome measures introduced above: Persuasion Susceptibility (PS) and Directional Steering (DS). PS captures reversibility under conversational challenge, while DS captures the nonnegative counterbalanced steering component toward the prompt-targeted response.

ERS combines these quantities as
\[
ERS = 1 - (\alpha PS + \beta DS),
\]
where $\alpha$ and $\beta$ are non-negative weights that sum to 1. In our experiments, we use $\alpha=\beta=0.5$, and report sensitivity analysis with alternative settings.

This formulation of ERS helps in separating general reversibility from target-directed steering. In the anti-baseline challenge protocol, the persuasive prompt targets the response opposite the baseline judgment, so apparent target alignment can coincide with decision reversal. The counterbalanced DS component separates this effect from genuine target-directed steering by comparing target alignment under persuasion against target alignment under neutral reconsideration.

\subsection{Ranking Analysis}

To assess system-level effects, we convert pairwise judgments into global rankings using the Bradley-Terry model. We then compare rankings before and after persuasion using Kendall's $\tau$. This analysis quantifies how local decision changes propagate to benchmark-level outcomes, affecting comparative model evaluation.

\subsection{Statistical Analysis}

We employ multiple statistical methods to validate our findings. McNemar's test compares paired outcomes between control and intervention conditions, with effect sizes reported using Cohen's $h$. We also fit generalized estimating equation (GEE) linear probability models (Gaussian family) with an exchangeable correlation structure, clustering at the prompt level to account for repeated measurements. A linear probability 
specification is used because quasi-complete separation in the binary outcome due to near-zero control flip rates versus high persuasion flip rates prevents convergence of logistic (binomial) specifications. Under this specification, coefficients are directly interpretable as changes in the probability of a decision flip. These models include covariates like persuasion type, judge model, dataset, baseline confidence, and response order.

\subsection{Controls and Validity}

We incorporate several controls to ensure internal validity. Repeated baseline evaluations quantify intrinsic instability. Neutral re-prompting isolates the effect of conversational context. Template paraphrasing verifies robustness to prompt wording. Bias analyses examine potential position and length effects. These controls support a causal interpretation in which observed decision changes are due to persuasive content rather than confounding factors.

\section{Results and Discussion}

We organize the results around three central findings. First, judges are highly stable under repeated and neutral evaluation, but stable judgments become reversible under conversational challenge. Second, anti-baseline challenge effects can propagate to human alignment and benchmark rankings, while a counterbalanced audit separates reversibility from target-directed steering. Third, common reliability signals, including confidence and revised justifications, fail to identify vulnerable judgments. Together, these results show that LLM-as-judge evaluation can be stable in form while remaining manipulable in interaction.

\subsection{Stable but Not Robust: Evaluation Breaks Under Challenge}

Intrinsic instability is near zero under repeated and neutral evaluation. Under repeated baseline evaluation, only 2 of 200 decisions change (1.0\%), and under neutral re-prompting, 0 of 200 decisions change. This confirms that LLM-based evaluation is highly stable in the absence of targeted persuasion. In contrast, conversational challenge produces substantial decision reversals. In the anti-baseline challenge protocol, 49\% of evaluations flip across persuasion conditions, with authority-based prompts reaching 74\%. In the counterbalanced target audit, where targets are assigned independently of the baseline judgment, reversibility remains present but smaller: persuasion susceptibility is $PS=0.194$, with authority again the strongest intervention (31.7\% flip rate). Thus, both protocols support the central claim that neutral stability does not imply robustness to conversational challenge. Figure~3 shows this transition from stability to manipulability: challenge sharply increases decision reversals while reducing human alignment in the stress-test setting.

\subsection{Challenge-Aligned Reversibility Under Persuasion}

In the anti-baseline challenge protocol, nearly half of post-challenge decisions align with the response suggested by the challenge: approximately 49\% select the response opposite the baseline judgment. This shows that conversational prompts can systematically influence evaluation outcomes rather than merely inducing random variation. However, because the anti-baseline challenge prompt targets the response opposite the initial decision, target alignment and decision reversal coincide in this protocol. We therefore interpret this result as challenge-aligned reversibility rather than standalone evidence of arbitrary target-directed steering. A counterbalanced target-validation protocol separates these effects: persuasion still increases reversibility ($PS=0.194$), especially under authority framing, but does not produce net target-directed steering beyond neutral reconsideration ($DS_{\mathrm{signed}}=-0.018$; clipped $DS=0$).

These results indicate that evaluation decisions are influenced by conversational framing, with the dominant failure mode being post-decision reversibility rather than unrestricted directional control.

\begin{figure}[t]
\centering
\includegraphics[width=\columnwidth]{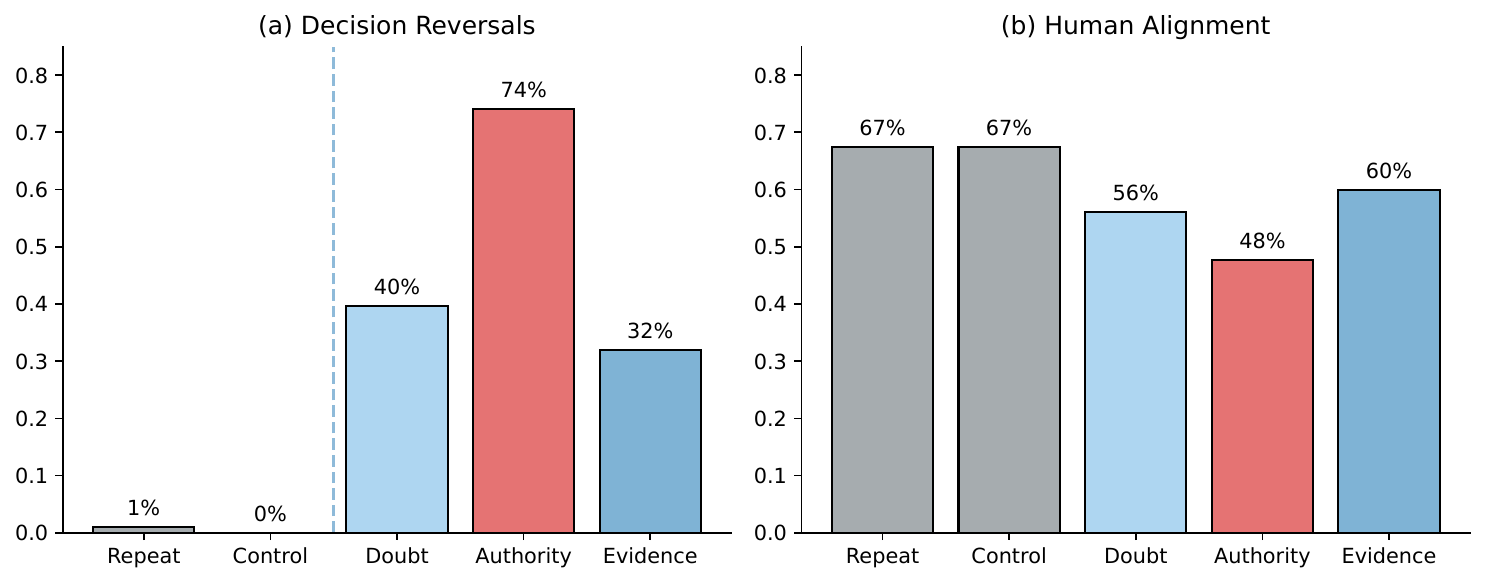}
\caption{\textbf{Conversational challenge induces reversals and degrades alignment.}
Flip rates increase sharply under persuasion (49\% overall; 74\% under authority), while agreement with human preferences declines 67\% to 48\%. The absence of flips under neutral control confirms that these effects are causally driven by interaction.}
\label{fig:causal}
\end{figure}

\subsection{Persuasion Degrades Human Alignment}
Under the anti-baseline challenge protocol, conversational challenge reduces agreement between LLM judges and human preferences. Alignment remains stable at 67\% under baseline and neutral conditions, but drops to 48\% under authority-based persuasion, a decrease of 19.8 percentage points. A majority of labeled reversals (64\%) are harmful, moving decisions away from human-preferred outcomes rather than correcting errors. While incorrect baseline decisions are more likely to be corrected (54\%), correct decisions also flip at a high rate (45\%), and the majority of baseline decisions are already correct (68\%). As a result, 64\% of labeled flips move away from human preferences.

The counterbalanced validation shows a more modest alignment effect. Baseline alignment is $63.7\%$, neutral alignment remains $63.7\%$, and the worst-case authority condition reaches $60.5\%$, a drop of 3.3 percentage points. Doubt and evidence conditions do not reduce alignment in this audit. We therefore interpret alignment degradation as strongest under the anti-baseline challenge and more limited under counterbalanced targeting.

\subsection{Conversational Challenge Destabilizes Benchmark Rankings}

Conversational challenge can propagate from local preference changes to benchmark-level rankings. Under the anti-baseline challenge protocol, Kendall's $\tau$ drops to 0.50, with 6 of 8 ranked entries changing position. Two entries, model\_A and model\_B, are aggregate AlpacaEval labels without generator metadata; among the six named candidate systems, the ranking effect is smaller. Figure~4 shows that most movement occurs among mid-ranked models while the top-ranked system remains stable.

The counterbalanced real-system validation shows stable pooled rankings across the six named systems ($\tau=1.00$, 0/6 changed; bootstrap 95\% CI $[0.60,1.00]$), but per-type analyses reveal modest drift: doubt and authority each yield $\tau=0.87$ with 2/6 systems changing rank, while evidence yields $\tau=0.73$ with 3/6 changing rank. Thus, ranking instability is strongest under the anti-baseline challenge setting, while the counterbalanced audit shows pooled stability with condition-specific sensitivity.

\begin{figure}[t]
\centering
\includegraphics[width=\columnwidth]{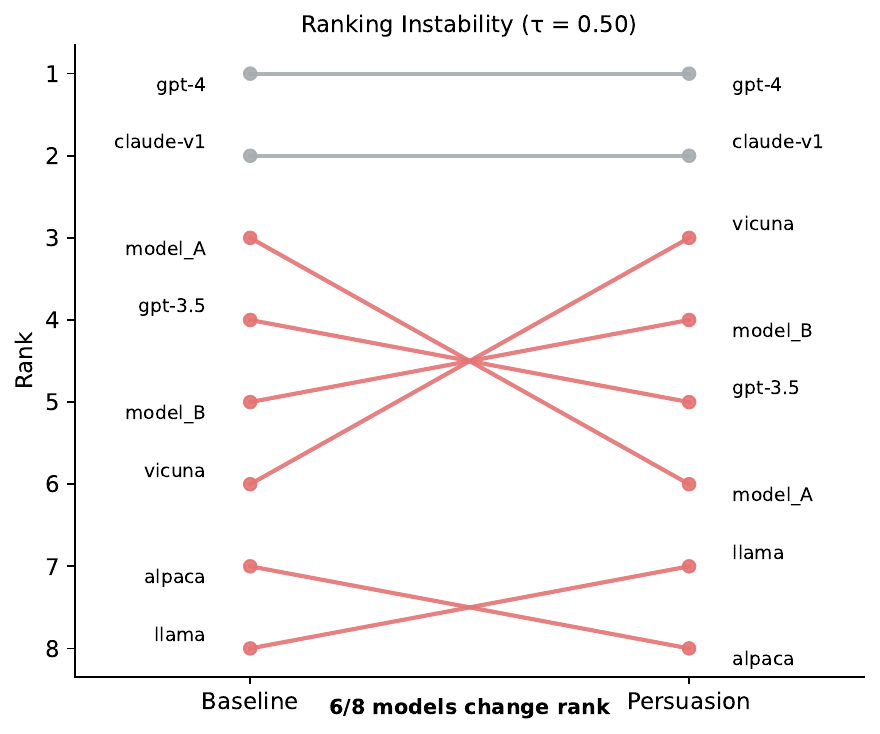}
\caption{\textbf{Ranking sensitivity under post-decision challenge.}
Under the anti-baseline challenge protocol, rankings derived from LLM judgments shift under persuasion (Kendall's $\tau=0.50$), with 6 of 8 ranked entries changing position.}
\label{fig:ranking}
\end{figure}

\subsection{High Confidence Does Not Imply Robustness}
\begin{figure}[!t]
\centering
\includegraphics[width=0.9\columnwidth]{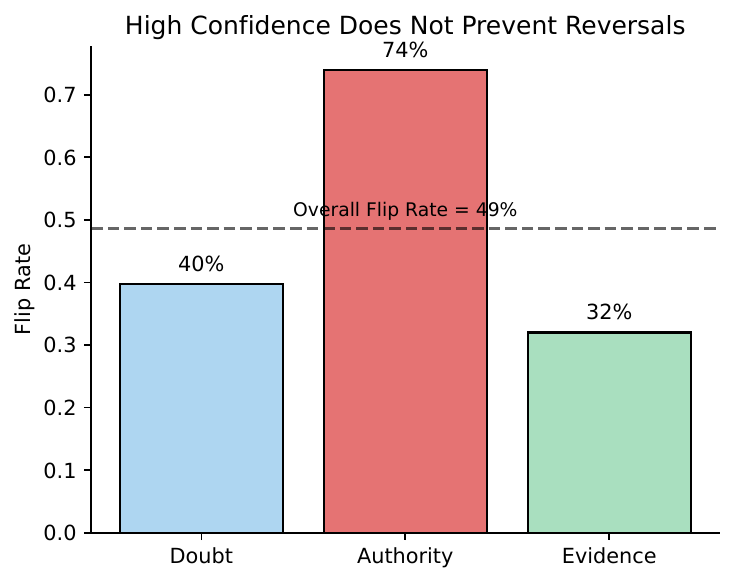}
\caption{\textbf{Confidence does not predict robustness.}
All evaluations fall within the high-confidence range, yet decisions still reverse at high rates under persuasion (49\%), indicating miscalibration, where confidence does not reliably predict robustness.}
\label{fig:confidence}
\end{figure}

All evaluations fall within the high-confidence range (70-100). This indicates that confidence scores cannot be used to identify or filter unreliable evaluations. Authority-based persuasion further highlights this issue. Although it produces the highest flip rate (74\%), it also leads to the largest decrease in confidence (-7.1), with 95\% of flips showing reduced confidence. This suggests that evaluation decisions are influenced more by social framing than by response quality. Figure~\ref{fig:confidence} shows that reversals occur despite uniformly high confidence, indicating a calibration failure where confidence unreliably predicts evaluation robustness.

\subsection{Decision Reversals Involve New Justifications}
Decision reversals are typically accompanied by newly generated justifications 
rather than revisions of prior reasoning. Under the anti-baseline challenge 
protocol, the average overlap between original and revised explanations is 
0.23, with 37\% of cases showing less than 20\% overlap, and confidence 
decreases on average ($-4.9$), with 81\% of reversals showing reduced 
confidence. The counterbalanced validation shows a similar pattern: overlap 
of 0.232, with 42\% of cases below 20\%, and an average confidence drop of 
$-6.6$ across 116 flipped cases. These patterns indicate post hoc 
rationalization after reversal rather than explicit error correction.

\subsection{Ambiguity Amplifies Vulnerability}

Evaluation difficulty strongly modulates susceptibility to persuasion under the stress-test protocol. For cases where judges agree at baseline (83 pairs), the flip rate is 43\%, but where judges initially disagree (17 pairs), the flip rate rises to 75\% ,a 1.7$\times$ increase. This is concerning because ambiguous cases are often the most informative for benchmarking. The evaluations that matter most are also the least robust.

\subsection{Multi-Step Challenges Produce Non-Monotonic Reversibility}

Multi-step persuasion shows that susceptibility does not simply accumulate with additional interaction. In the counterbalanced run, 10.2\% of decisions differ from baseline after the first doubt-based challenge. This proportion rises to 39.0\% after the authority challenge, then falls to 18.6\% after the subsequent evidence-based challenge. Across the full multi-step trajectory, 27 of 59 decisions flip at least once, and the average number of challenges required to first induce a flip is 1.89. These results indicate that post-decision influence is dynamic rather than monotonic. Authority framing produces the strongest intermediate disruption, but later interaction can partially restore the original judgment. Thus, multi-step challenge exposes reversible evaluation behavior: judgments may move away from and back toward the baseline depending on the framing of subsequent interaction.

\begin{table*}[t]
\centering
\scriptsize
\setlength{\tabcolsep}{3.2pt}
\renewcommand{\arraystretch}{1.15}
\begin{tabularx}{\textwidth}{
@{}
>{\raggedright\arraybackslash}p{0.15\textwidth}
>{\centering\arraybackslash}p{0.11\textwidth}
>{\centering\arraybackslash}p{0.15\textwidth}
>{\centering\arraybackslash}p{0.22\textwidth}
>{\raggedright\arraybackslash}X
@{}}
\toprule
\textbf{Property} 
& \textbf{Metric} 
& \textbf{Baseline / Control} 
& \textbf{Persuasion / Audit} 
& \textbf{Key Finding} \\
\midrule

\textbf{Stability} 
& Flip Rate 
& \makecell{1.0\% repeat\\0\% neutral}
& \makecell{\textbf{49\%}\\anti-baseline reversal}
& Stable under repetition, but highly reversible under challenge \\

\textbf{Persuasion Susceptibility} 
& PS 
& 0\% neutral flips
& \textbf{19.4\% counterbalanced validation}
& Persuasion induces reversals beyond neutral reconsideration \\

\textbf{Target Alignment / Steering} 
& \makecell{Target Align.\\DS}
& \makecell{51.3\%\\neutral target align.}
& \makecell{\textbf{49\%} anti-baseline target align.\\49.4\% validation target align.\\$DS_{\rm signed}=-0.018$; $DS=0$}
&Anti-baseline target alignment is high, but counterbalancing shows no net target-directed steering \\

\textbf{Human Alignment} 
& Agreement 
& \makecell[tc]{67\% baseline\\63.7\% validation baseline}
& \makecell[tc]{\textbf{48\%} authority challenge\\60.5\% authority validation}
& Alignment drops strongly under anti-baseline challenge and modestly under counterbalanced validation \\

\textbf{Harmfulness} 
& Harmful Flip Ratio 
& -- 
& \textbf{64\%}
& Majority of labeled stress-test reversals degrade evaluation quality \\

\textbf{Ranking Stability} 
& Kendall $\tau$ 
& 1.00 
& \makecell{\textbf{0.50} anti-baseline\\1.00 pooled validation\\0.73 per-type min}
& Rankings shift under anti-baseline challenge; validation shows pooled stability with condition-specific drift \\

\textbf{Confidence} 
& Mean Confidence 
& 89\% 
& 82\% 
& High confidence despite reversals; confidence remains miscalibrated \\

\textbf{Authority Effect} 
& Flip Rate 
& -- 
& \makecell{\textbf{74\%} anti-baseline\\31.7\% validation}
& Authority is the strongest destabilizing challenge \\

\textbf{Reasoning} 
& Overlap 
& -- 
& \makecell{0.23 anti-baseline\\0.232 validation}
& New justifications after reversal show low overlap \\

\textbf{Robustness} 
& ERS 
& -- 
& \makecell{\textbf{0.51} anti-baseline\\0.903 validation}
& Anti-baseline vulnerability is high; corrected ERS localizes failure to reversibility \\

\bottomrule
\end{tabularx}
\caption{\textbf{Summary of evaluation robustness under conversational challenge.}
Evaluation is stable under repetition and neutral reconsideration but reversible under post-decision challenge. The counterbalanced target-validation protocol separates this reversibility from target-directed steering, showing that the dominant failure mode is post-decision reversibility rather than unrestricted directional control.}
\label{tab:summary}
\end{table*}

\subsection{ERS Reveals a Reversibility-Dominated Failure Mode}

ERS summarizes the robustness findings across both the anti-baseline challenge protocol and the counterbalanced target-validation protocol. Under the targeted post-decision challenge, ERS remains low at approximately 0.51, placing judges in a susceptible setting despite high repeat and neutral stability. The counterbalanced validation clarifies the source of this susceptibility. Persuasion susceptibility remains nonzero ($PS=0.194$), and authority remains the strongest intervention, but the directional component is not positive ($DS_{\mathrm{signed}}=-0.018$; clipped $DS=0$). As a result, the counterbalanced ERS is higher ($ERS=0.903$), indicating that the dominant failure mode is reversibility rather than net arbitrary target steering.

Thus, repeatability alone would classify these judges as reliable, yet challenge-based evaluation reveals substantial susceptibility under stress testing and persistent reversal vulnerability under counterbalanced validation. Table~\ref{tab:summary} summarizes these findings. We report two complementary analyses: an anti-baseline stress test, which measures whether stable decisions can be overturned, and a counterbalanced target-validation audit, which separates reversibility from net target-directed steering.

\subsection{Implications for Evaluation Practice}

These results highlight a gap in current evaluation methodology. Stability under repeated evaluation does not imply robustness under interaction, and existing mitigation strategies, including confidence filtering and multi-judge aggregation, fail to address this vulnerability. Evaluation pipelines should explicitly account for adversarial interaction, for example, by incorporating challenge-based diagnostics, constraining post-decision interaction, or reporting robustness metrics such as ERS. More broadly, these findings highlight a tension between adaptability and reliability.

\section{Conclusion}

This work revisits a central assumption in LLM-as-judge evaluation: that evaluation outcomes remain stable once produced. Our findings show that this assumption does not fully hold under interaction. Judges are highly stable under repeated and neutral conditions, but their decisions become reversible under conversational challenge. In the anti-baseline challenge protocol, judges often abandon stable baseline decisions when prompted to reconsider in favor of the alternative response.

A counterbalanced target audit clarifies the nature of this vulnerability. Persuasion-induced reversibility remains present, especially under authority framing, but the audit does not show net arbitrary target-directed steering. This distinction is central: anti-baseline stress tests reveal whether judgments can be overturned, while counterbalanced audits test whether those changes reflect genuine target-directed control.

This work highlights a fundamental tension in LLM-based evaluation between adaptability and reliability. The same capacity that enables models to revise and improve their outputs also makes them vulnerable as evaluators. These results show that post-decision robustness should be treated as a separate dimension of LLM-as-judge evaluation. Evaluation pipelines should report not only static agreement or repeatability, but also susceptibility to post-decision interaction. ERS provides one way to quantify this vulnerability by separating reversal susceptibility from counterbalanced directional steering.

\section{Limitations and Future Work}

\subsection{Limitations}

This study focuses on pairwise evaluation in instruction-following benchmarks using two judge models, GPT-4o and GPT-4o-mini. These models are representative of widely used LLM evaluators, but susceptibility to conversational persuasion may differ across other architectures or future model versions. Similarly, while the experiments cover two standard benchmarks (MT-Bench and AlpacaEval), the extent to which these effects generalize to other domains, tasks, or evaluation settings remains to be explored.

The experiments use a controlled conversational setup in which the judge is asked to reconsider its decision after a targeted post-decision challenge. This design is useful for isolating the causal effect of interaction, but real-world evaluation pipelines may include additional safeguards, such as judge aggregation, rubric-locked evaluation, independent adjudication, or restricted post-decision access. We leave open how these safeguards interact with post-decision influence, and whether they reduce harmful reversibility while preserving legitimate error correction.

The dataset size is relatively modest (100 pairs), but each pair is evaluated under multiple controlled conditions (1,440 total evaluations), providing strong statistical power. The effects observed are large, consistent across models and datasets, and statistically significant. While larger-scale validation would be valuable, the consistency of these results suggests that the qualitative findings are likely to generalize.

A further consideration is that the anti-baseline challenge protocol is intentionally diagnostic rather than merely observational. By targeting the response opposite the baseline judgment, it provides a direct test of post-decision overturnability: whether a judgment that is stable under repetition and neutral reconsideration remains reliable when challenged to favor the alternative response. This captures a practically important robustness failure mode for interactive evaluation systems. The counterbalanced target-validation protocol then complements this test by separating persuasion-induced reversibility from net target-directed steering. Together, the two protocols provide a framework for measuring interactional robustness while distinguishing reversibility, directional susceptibility, and harmful versus beneficial revision.

Finally, our analysis characterizes the behavioral manifestation of post-decision manipulability rather than its underlying mechanisms. The observed patterns, including authority sensitivity, confidence miscalibration, low-overlap revised justifications, and harmful reversals, suggest connections to broader forms of conversational compliance and sycophancy. However, identifying the relative roles of instruction tuning, preference optimization, evaluator prompting, and conversational context requires targeted mechanistic study.

\subsection{Future Work}

This work motivates a broader research agenda on interaction-robust evaluation. Future evaluation protocols should measure not only whether an LLM judge agrees with human preferences in a static setting, but also whether its judgments remain reliable when the evaluation process becomes interactive. In particular, robust LLM-as-judge evaluation should report three complementary quantities: repeat stability, anti-baseline reversibility, and counterbalanced target-directed steering.

One important direction is cross-architecture robustness evaluation. The present study establishes post-decision manipulability for two GPT-4o-family judges, but leaves open whether this vulnerability is model-specific or structural to LLM-as-judge evaluation. Future work should apply the same within-instance protocol across open-weight models, proprietary systems, evaluator-specialized models, reward models, and ensemble-based judging systems. Such comparisons would clarify whether interaction robustness depends primarily on model family, scale, training procedure, evaluator prompting, or aggregation strategy.

A complementary direction is to characterize the prevalence and boundary conditions of post-decision manipulability. Larger and more diverse evaluations should vary task type, domain, language, modality, response length, evaluation difficulty, baseline confidence, and baseline judge disagreement. This would help identify where post-decision influence is most consequential and whether susceptibility follows predictable patterns across benchmark design choices. Ambiguous or high-disagreement cases are especially important, because these are often the cases where evaluation decisions matter most.

Future work should also standardize target-validation protocols for studying judge manipulation. Anti-baseline challenges are useful for testing whether stable decisions can be overturned, but counterbalanced target assignment is necessary for estimating whether interventions produce net target-directed steering. We therefore recommend treating counterbalanced target validation as a standard component of interaction-robust evaluation, especially when claims involve directional influence or manipulability.

Mitigation is another important direction. The goal should not be to prevent judges from ever revising their decisions, since revision can reflect legitimate error correction. Rather, the challenge is to distinguish warranted revision from conversationally induced preference change. Promising approaches include heterogeneous judge panels, rubric-anchored reconsideration, structured revision protocols that require judges to identify a substantive evaluation error before changing a preference, separation of initial and revised judgments, and training or prompting methods aimed at reducing sycophancy and authority sensitivity. Future work should evaluate whether these interventions reduce harmful reversals while preserving beneficial reconsideration.

Finally, future work should investigate the mechanisms that give rise to post-decision manipulability. A useful next step is to compare base, instruction-tuned, preference-tuned, and evaluator-specialized models under the same anti-baseline and counterbalanced target-validation protocols. Such studies could clarify whether susceptibility arises primarily from general conversational compliance, alignment training, evaluator prompting, or their interaction. Understanding these mechanisms is necessary for moving from diagnostic robustness evaluation toward judge designs that are both adaptive and reliable.

\section{Ethical Considerations}

This study uses existing, well-established benchmarks such as MT-Bench and AlpacaEval. No new human-subjects data were collected, and no private user data were used. The work identifies a reliability risk in LLM-as-judge evaluation: post-decision interaction can change judge preferences even when the evaluated responses are held fixed. Because this behavior has implications for the integrity of automated evaluation pipelines, care should be taken when allowing post-decision interaction with judge models, and additional robustness checks are needed before such systems are used in consequential evaluation settings. The intended use of the work is to support more robust evaluation practice, including clearer interaction boundaries, challenge-based robustness tests, and reporting of judge susceptibility under conversational challenge.

\bibliography{references}

\end{document}